\documentclass{article}[15pt]

\usepackage{amsmath,epsfig,color,calc,rotating,graphics,ulem}

\usepackage[pagewise]{lineno}

\usepackage{natbib}
\bibpunct{(}{)}{;}{a}{}{,}

\usepackage{url}
\usepackage{float}

\begin{document}

\title{
Quadratic Term Correction on Heaps' Law  \\
\vspace{0.2in}
\author{
Oscar Fontanelli$^1$ and Wentian Li$^{2,3}$ \\
{\small \sl  1. Facultad Latinoamericana de Ciencias Sociales M\'{e}xico, CD M\'{e}xico, M\'{e}xico }\\
{\small \sl 2. Department of Applied Mathematics and Statistics, Stony Brook University, 
Stony Brook, NY, USA } \\ 
{\small \sl  3. The Robert S. Boas Center for Genomics and Human Genetics}\\
{\small \sl  The Feinstein Institutes for Medical Research,  Northwell Health, Manhasset, NY, USA}\\
}
\date{May 24, 2026}
}
\maketitle  
\markboth{\sl Fontanelli and Li}{\sl Frontanelli and Li}

\normalsize

\vspace{0.5in}

\begin{center}
Abstract
\end{center}

One of the best known linguistic, bibliometric, informetric, system sciences' laws, 
the Heaps' or Herdan's law, characterizes the type vs. token relation
by a power-law function, which is concave in linear-linear scale but a straight
line in log-log scale. However, it has been observed that even in log-log
scale, the token-type curve is still slightly concave, invalidating the power-law
relation. At the next-order approximation, we have shown, by twenty English
novels or writings (some are translated from another language to English), 
that quadratic functions in log-log scale fit the token-type data perfectly.
Regression analyses of log(type)-log(token) data with both
a linear and quadratic term consistently lead to a linear coefficient of
slightly larger than 1, and a quadratic coefficient around $-0.02$.
Using the ``random drawing colored ball from the bag with replacement" 
model, we have shown that the curvature of the log-log scale is identical
to a ``pseudo-variance" which is negative. 
Although a pseudo-variance calculation may encounter numeric instability
when the number of tokens is large, due to the large values of pseudo-weights,
this formalism  provides a rough estimation of the curvature when the number of tokens
is small.

\newpage

\large

\section{Introduction}

\indent

As languages can be considered as complex systems \citep{smith,larsen,kret,drozdz},
linguistic laws may be manifestations of statistical \citep{altmann}, 
or physical \citep{torre}, or evolutionary \citep{lieberman} pattern in complex systems.
Linguistic laws often appear in non-linguistic situations, such as 
bibliometrics \citep{bailon}, information retrieval \citep{petersen},
biology \citep{semple}, and complex systems \citep{stanisz}.

One of the linguistic laws, the Heaps' law \citep{heaps} or Herdan's law 
\citep{tt-herdan} describes a linguistic
trend that the number of dictionary words used in a text,
or, vocabulary size ($V$), is a power-law function of
the text length measured by the number of token words ($T$): $V = c T^\alpha$, 
or $V \propto  T^\alpha$,
where the exponent $\alpha < 1$. The word ``words" is used twice in the above
sentence: in order to distinguish the two different meanings, the first use
is also called ``type" (unique, distinct, dictionary words), and the second use is called
``token" (might be repeated, usually not unique, only a measure of number of appearance).
Even these explanations need to specified, for example, whether ``work", ``works", ``worked"
are considered to be three types or one type.
The inequality $\alpha < 1$ is strict: if $\alpha=1$, then the number of types is
proportional to the number of tokens, which is unsupported by the data.

Linguistic laws are rarely considered or proven to be exact. Zipf's law describes the
number of token for a specific type, when ranked from the most frequently
used to the rarest used one, as a power-law function of rank ($r$): $T_r \propto 1/r^\beta$
where $\beta \approx 1$ \citep{zipf35}. Although Zipf's law is rather good
in summarizing the word usage pattern in many different languages, in 
particular considering the simplicity of the exponent being 1, the 
fitting of real linguistic data is not perfect \citep{wli-entropy,piantodosi, moreno,mehri}.
Furthermore, if some words are removed, such as stopwords,
the rank-frequency distribution of
the remaining word tokens no longer follows the Zipf's law \citep{wli-stop}.

Heaps-Herdan's law suffers from a same fate. 
If one piece of work (e.g. one novel)
contributes one point (i.e., the total number of word types and word tokens -- length
of the text) in the token-type plot, we would expect a noisy scattering of points,
simply because different authors or difference source of texts may have their distinct
word-type usage. This is the first version of Heaps' law  described in \citep{chacoma}.  

Checking the validity of Heaps' law for a given piece of work by a specific author
is relatively simple (this is the third variant of Heaps' law in \citep{chacoma}): 
one ``window" of text contributes one point in the token-type plot, and all other
windows are from this piece of work. This version promises a smoother curve, but
there are still sub-versions on how windows are chosen. 
One sub-version is by partitioning the whole text into evenly sized windows,
and all windows thus constructed are included in the token-type plot. This way,
at a given $x$ value (given token count, or given window size), there is a
scattering of dots in token-type plot. Another sub-version is to always choose
windows starting from the beginning -- this way, there is only one dot in the
token-type plot when $x$ is fixed.  

Keeping in mind of these versions and sub-versions of token-type plots,
there were several studies pointing out that token-type relation does not
follow a straight line in log-log scale in all $x$ ranges 
\citep{bernhardsson,lv2013,font}. In particular,
the scatter plots are downwardly concave. It implies that the power-law 
relationship itself in log-log plot is not exactly correct.  If the slope 
in log-log scale is calculated locally, its value is changing gradually.
In this case that one parameter in Heaps' law $\alpha$ is not enough to characterize 
the full scope of the relationship, then the power-law assumption is no longer correct,
or not exactly correct.  Previously, we have observed a similar bending 
in a different context, for the DNA words constructed by the chromosomal
regions that code protein domains \citep{wli24}. 

In this paper, we will examine the deviation from power-law function for token-type 
plot in human language texts and propose alternative fitting functions. 
To show a systematic deviation from the linear trend, we will show that for
local linear regressions, the slope values tend to decrease with larger window sizes.
A traditional solution in dealing with deviation from a linear
relationship is to add a correction term. As in Taylor expansion, the next
correction after the linear term is the quadratic term, i.e.,
\begin{equation}
\label{eq-q}
\log(V) = c_0 + \alpha \log(T) + \beta \left( \log(T) \right)^2
\end{equation}
or, equivalently ($c\equiv e^{c_0}$),
\begin{equation}
\label{eq-eff-exp}
V = c T^{\alpha + \beta \log(T)}
\end{equation}
On surface, it is still a Heaps' law (power-law), but the effective exponent 
$\alpha_{eff}= \alpha +\beta \log(T)$ is not a constant, but changes with the 
number of tokens $T$. In particular, if $\beta <0$, the
effective exponent decreases with $T$. 

The token-type relationship, at least for the last sub-version
we mentioned, can be modelled by a random
ball-drawing model. Suppose there are total $T_{tot}$ number of balls in a bag
(or an urn)
with $V_{tot}$ different colors. Randomly drawing $T$ balls from the bag,
how many different colors does one expect to see? This question was asked
in ecological studies, on how many different species one expect to see
after obtaining certain number of samples. A solution was proposed as early as 1971
\citep{hurlbert} (see also \citep{heck,milicka09,font,davis,chacoma,debowski}, and
Sec 2.2 of \citep{baayen}) for either sampling without replacement:
\begin{equation}
\label{eq-wo-replace}
E[V(T)]= V_{tot} - \sum_{i=1}^{V_{tot}} \frac{ \left( 
\begin{matrix} T_{tot} -T_i \\ T  \end{matrix}\right)}
{\left(\begin{matrix} T_{tot} \\ T \end{matrix} \right)}
\end{equation}
or sampling with replacement:
\begin{equation}
\label{eq-w-replace}
E[V(T)]= V_{tot} - \sum_{i=1}^{V_{tot}}
\left( 1-\frac{T_i}{T_{tot}}\right)^T
\end{equation}
where $T_i$ is the number of color-$i$ balls in the bag, and $E[V(T)]$ is
the expected/average number of colors when sampling $T$ balls. The derivation
of these formula is based on this simple argument that
Prob(see color $i$)=Prob(color-$i$ appear $\ge$1 times)=1-Prob(color-$i$ appear 0 times),
and summing over all colors is the expected number of colors.
Note that these formula require knowing the value of $T_{tot}, V_{tot}$, as well
as total number of balls with individual colors $T_i$.

For a small window, due to the correlation among words
used, the random model may not be appropriate. However, the last point or dot
on the token-type plot, i.e., ($T_{tot}, V_{tot}$), is fixed, and
random model should be close to be correct when high proportion of tokens are
sampled. The leading term approximation of Eq.(\ref{eq-wo-replace})
can be found in \citep{font}, and that of Eq.(\ref{eq-w-replace})
can be found in \citep{boytsov}. The focus on the leading term is
to derive a relationship between the Heaps' law exponent and Zipf's law
exponent.

Here, we tackle the problem of deviation from the power-law in token-type
plot from two angles. The first is a more careful analysis of token-type
plots on texts in 20 books, as well as in different sub-versions. 
Due to the scattering of dots in $y$ direction
in some versions, and uneven distribution of dots in the $x$ direction
after the log-transformation, regression analysis may lead to different results.
We would provide recommendation on which choices are ``better". Then, we
will turn attention to the random-ball-drawing model and carry out approximation
focusing on the second-order term. It turns out that the curvature
in the log-log scale, which is the second derivative of log(type) over log(token),
can be derived as  a ``pseudo variance". The difference between a pseudo
variance and a real variance is that the ``weight" in the former can contain
negative values. Both analytic check on formula with fewer number of tokens 
and numeric calculation show that this pseudo variance is negative. Therefore, the
analytic results and real text analysis converge to the same conclusion.

The paper is organized as follows: The results are split into
two sections, one on fitting real English texts with quadratic regression
in log-log scale of token-type plot, another on exploring the properties of the 
random ball drawing model.  Each result section consists of these subsections:
Sec~\ref{sec:war-peace} introduces the issue by examining the texts from {\sl War and Peace},
showing that quadratic regression is better than linear regression in log-log scale
of token-type plot, in two different ways of plotting type vs. token, and with two
different measures of goodness of data fitting;
Sec~\ref{sec:books} extends that result to texts from  19 other books downloaded
from the Project Gutenberg (thus total 20 books);
Sec~\ref{sec:pseudoV} provides an analytic result from the random ball drawing model
with replacement, that the curvature of the log-log scale token-type plot is of
the form of ``pseudo-variance";
Sec~\ref{sec:negative} shows by both analytic (from three tokens) and numeric
calculations that the pseudo-variance discussed in the previous subsection is negative;
Sec~\ref{sec:zipfexp} shows that if the word type distribution follows a Zipf's law
(inverse-power-law) with an exponent, the quadratic regression coefficient decreases
with an increase of that exponent.

\section{Results from analysis of English texts}

\subsection{ Deviation from the Heaps' law -- the example of War and Peace}
\label{sec:war-peace}

\indent

Fig.\ref{fig1} shows a token-type scatter plot for Leo Tolstoy's ``War and Peace" in log-log
scale. 
It is the version of using single piece of work, and the sub-version
of using all windows that evenly partition the whole book with $2^k$ number of windows.
The rightmost point correspond to the whole text with more than half
a million tokens. Other dots in the plot correspond to tokens in non-overlapping
windows throughout the text. For example, the smallest window size is 100 tokens;
we take the first 100 tokens, count the number of types, then move to the next
(non-overlapping) 100 tokens, count types, etc. To help the visual impression
of these dots, noise is added in their $x$-values.

Regression analyses were carried out in two different ways: (1) to fit 
all dots (grey dots in Fig.\ref{fig1}) by a regression line, and (2) only
use one value per $x$-value, e.g. median for all dots with the same $x$-value 
(blue crosses in in Fig.\ref{fig1}), for regression.  Due to their 
larger number, dots with smaller windows will weight more on the first regression, whereas the
second regression method would treat all window sizes equally.

If the Heaps' law is correct, the power-law exponent would be the regression
coefficient $\alpha$ in $\log(V)= c_0 + \alpha \log(T)$,
where $T$ is the number of tokens, $V$ the number of types, in a window. 
The first regression (using all dots) leads to $\alpha=0.756$ (pink solid line 
in Fig.\ref{fig1}), 
and the second regression method (using only medians) leads to $\alpha=0.662$ 
(blue solid line in Fig.\ref{fig1}).
It is clear from visual inspection that the second regression method captures the trend
in the token-type relationship better than the first regression. 
However, the second regression line still has systematic bias (the line is on top of the data 
at the two ends, but below the data in the middle).
The conclusion is that the functional form in the Heaps' law, i.e., 
the power-law function, is not strictly correct.

A natural step to improve the data fitting performance is to increase the number
of parameters by one \citep{frappat,wli-entropy,wli-letter,oscar-lavalette,oscar-brf}, 
such as the  quadratic equation
of Eq.(\ref{eq-q}). The two regression lines by the above mentioned two methods
(using all dots and using the median) by of Eq.(\ref{eq-q}) are shown in
Fig.\ref{fig1} (pink and blue dashed lines). Looking more carefully, it can
be seen that the quadratic regression using all windows (blue dashed line) fails
to fit the last few points. The winner of regressions tried so far 
is the quadratic regression over medians (the  blue 
dashed line that fits blue crosses).

To quantify how much better the quadratic function is in fitting the
(log)type-(log)token relationship better than linear functions (i.e., Heaps' law), 
the main issue is about how to penalize the extra parameter used.
The adjusted $R$-square \citep{ezekiel} penalizes models with more parameters by making 
the $R$-square (percentage of variance explained by the regression) smaller: 
\begin{equation}
\label{eq-r2adj}
R^2_{adj}= 1- \frac{(1-R^2)(n-1)}{(n-p-1)}, 
\end{equation}
here $n$ is the number of points to fit 
by regression, $p$ is the number of parameters in the regression.
($R$-square itself is defined as $R^2=1- RSS/TSS$ where RSS is the residue sum of
square and TSS is the total sum of square.) 
The Akaike information criterion (AIC) \citep{aic} is another model selection measure,
defined as 
\begin{equation}
\label{eq-aic}
AIC=n \log \left( \frac{RSS}{n} \right) +2p
\end{equation}
in the context of regression. The lower
the RSS, the better the regression fitting, whereas more parameter (larger $p$)
will penalize it. It makes AIC a compromise between data fitting and model complexity.

Table \ref{table1} shows that quadratic regressions are better than linear regressions,
for using both all grey dots in Fig.\ref{fig1}  and only the (blue crosses) median values
in Fig.\ref{fig1}, by having higher adjusted $R^2$, and by having lower AIC values.
 
\begin{table}[H]  
\begin{center}
\begin{tabular}{cc|cc}
\hline
regression & quantity &scatter (n = 10597) & median (n=12) \\
\hline
linear & $R^2_{adj}$ & 0.9898 & 0.9938 \\
& AIC  & -25148.98 & -8.51 \\
\hline
quadratic & $R^2_{adj}$ & 0.9911 & 0.9999 \\
 & AIC & -26577.83 & -59.28 \\
\hline
\end{tabular}
\end{center}
\caption{ \label{table1}
Comparing the regression performance in fitting $y=\log$(type) vs $x=\log$(token)
plot by linear and quadratic regressions, for the text in {\sl War and Peace}. 
There are two different ways (versions) to prepare the data to fit: using
all $n=10597$ moving windows, or, combining all windows with the same size
and replacing them by their mean (number of types). Both adjusted $R$-square
(the larger, the better the model) and AIC (the smaller, the better the model) are used.
In all cases, quadratic regression is a better model than the linear regression.
}
\end{table}

\subsection{Gradual decrease of local linear regression slope in 20 books}
\label{sec:books}

\noindent

We expand our examination of token-type plot from {\sl War and Peace}
to 19 other eBooks -- most of them were originally written in English while some
were translated to English from Latin, French and Spanish, and most are novels while
some are treatise on philosophy, natural history, economics, or other topics.
The number of tokens (length of the books in the unit of words) range from 23k
({\sl Discourse of Method} by Ren\'{e} Descartes) to 566k ({\sl War and Peace}).
Fig.\ref{fig2} shows the token-type plot in log-log scale for these 20 books.
The $x$-values (number of tokens per window) have an increment of 1 token,
ranging from a small window size all the way to the text length.  

The $y$-level at the same $x$-value indicates the vocabulary size at the same book 
length. The books with the least dictionary words (i.e. word type) used
include {\sl Alice's Adventure in Wonderland},
{\sl Wealth of Nation, Origin of Species, 
An Essay Concerning Humane Understanding, Pride and Prejudice}.
It is very likely that the authors of these books tried to communicate
with their readers (e.g. children, in the case of {\sl Alice's Adventure}) 
with simple languages. 

The book with the most word type is {\sl Ulysses}, which is consistent with its
reputation of innovation of styles and complexity. The book
{\sl Natural History of Pliny} (Vol.1) is encyclopedia like and it has a sudden jump
of word types between 10k and 100k. 
One explanation is that there are
inhomogeneities in the topics covered when the window sizes are expanded.
Drawing individual token-type plots of these 20 books, after the linear trend
removed (by linking the first and the last dot), one may still see local
kinks here and there; though all of them have a universal shape of upside-down parabola
(results not shown).

In order to show that these lines do bend downward, we
calculate the local slopes by focusing only on one region at the time,
spanning half a unit in the $\log_{10}(T)$ scale (or, from a $T$ value to
3.16T value). Local slope in log-log scale is also called elasticity 
\citep{marshall} or logarithmic slope  in the
original (token-type) function: $d \log(V)/d \log(T)$, mostly relevant
to econometric applications where log-log models are common.  

Fig.\ref{fig3} shows the local slopes for the twenty individual books.
Again, it is from sub-version of token-type plot where
the windows always start from the first token (same as Fig.\ref{fig2}).
For most books, the local slope starts high at smaller
window sizes, then gradually decreases with the window size,
with the exceptions of these books: {\sl An Essay Concerning Humane Understanding},
{\sl Natural History of Pliny}, and {\sl Ulysses}.
This proves that for the majority of works,
the log(type)-log(token) line curves downward.

Having shown that quadratic regression fits the {\sl War and Peace} 
log(type)-log(token) line better than linear regression, and other books
have the similar downward concave trend, here we intend
to find out typical features of $\alpha$ and $\beta$ values in Eq.(\ref{eq-q}).
For that we need to address another practical issue: should we use
the raw data which is not evenly distributed in the log scale of tokens, 
or should we use evenly distributed sample points in log-scale?  

In Table \ref{table2}, we show the quadratic regression results
using both approaches. On the left half of Table \ref{table2},
all windows are used, where the token count increases
from 1 to the book length (the smallest value is 23k tokens for 
{\sl Discourse on Method}, and the largest value is 566k tokens
 for {\sl War and Peace}). On the right half, we start from 
the 1.01 rounded to 1, the first data point, then continue at $k$ step
($k=2,3, \cdots$), from the round($1.01^k$)'th data point, etc. 
This way, windows are sampled evenly in the log scale of tokens. 
The number of sampled data points range from 646 to 968. 

\begin{table}[H]  
\begin{center}
\begin{tabular}{|c|c|ccc|cccc|}
\hline
title & text length& \multicolumn{3}{c|}{1-init window } & \multicolumn{4}{c|}{evenly sampling } \\
& & \multicolumn{3}{c|}{increments by 1 token} & \multicolumn{4}{c|}{in log scale} \\
\cline{3-9} 
 & =num-pts& $\alpha$ & $\beta$ & $R^2_{adj}$ & num-pts &  $\alpha$ & $\beta$ & $R^2_{adj}$\\
\hline
Discourse on Method & 23,037 &   1.22 & $-0.036$  & 0.9983 & 646 & 1.04 & $-0.022$  & 0.9979   \\
Alice's Adventure & 26,687 & 1.03 & $-0.025$ & 0.9997 &661 &  0.96 & $-0.02$& 0.9996\\
Hamlet & 31,966 & 	 0.96 &  $-0.015$ &  0.9996 & 679 &  0.92& $-0.013$ & 0.9994 \\
Beowulf & 36,171 &   1.21 & $-0.031$ & 0.9995  & 691&   1.11 & $-0.024$& 0.9996\\
Utopia & 42,249 &   1.02 & $-0.022$ & 0.9998 & 707 &  1.06& $-0.025$ & 0.9996 \\
Great Gatsby& 48,661 &   1.15 & $-0.028$ & 0.9988 &721 &   1.08 & $-0.023$  & 0.9998 \\
Frankenstein& 69,632 &   1.32 & $-0.039$ & 0.9991 & 757 &   1.09 & $-0.024$ & 0.9987\\
Natural History& 90,675 &  0.13 & 0.037  & 0.99 & 784 &  0.86& $-0.0059$& 0.9971 \\
Pride \& Prejudice& 122,971&   1.19 & $-0.032$ & 0.9984 & 814 &  0.97 & $-0.019$  & 0.9994 \\
Human Understanding& 143,672&   1.01 & $-0.023$ & 0.9951  &830 &  0.92 & $-0.019$ & 0.9985 \\
On Origin of Species & 151,077 &  1.11 & $-0.029$ & 0.9995 &835&   1.09 & $-0.028$ & 0.9996 \\
Dracula & 162,019 &   1.14 & $-0.028$ & 0.9985 		& 842 &  0.99& $-0.02$ & 0.9997\\
Leviathan & 211,782 &   1.09& $-0.024$& 0.9987 		& 869 & 0.91 & $-0.015$& 0.9996\\
Moby Dick & 212,510 &   1.09 &  $-0.022$ & 0.9990 	& 869 & 0.99 & $-0.016$ & 0.9997\\
Ulysses & 265,191 &  0.85 &  $-0.005$ & 0.9989  	& 892 & 0.89 & $-0.073$& 0.9996\\
Middlemarch& 320,171 &   1.17 & $-0.029$ & 0.9985 	& 910 &  1.11 & $-0.026$ & 0.9998\\
Bleak House& 357,408&  1.15 & $-0.028$  & 0.9983 	& 922 &  1.11 & $-0.026$ & 0.9994\\
Don Quijote &376,691 &  0.96 & $-0.016$ & 0.9993 	& 927&   1.03 & $-0.019$& 0.9998\\
Wealth of Nations& 380,995 &  0.92 & $-0.018$ & 0.9978 	& 928 &  1.03 & $-0.023$ & 0.9993\\
War and Peace& 566,051 &   1.07 & $-0.024$ & 0.9996 	& 968&  1.05 & $-0.023$& 0.9998\\
\hline
median & & 1.09& $-0.025$ & & & 1.03 & $-0.021$ & \\
\hline
\end{tabular}
\end{center}
\caption{ \label{table2}
Quadratic regression results on 20 texts (see Sec.\ref{sec:data}) in two versions. 
One version (on the left) considers each 1-init window as a data point, and
the total number of data points is the text length (total number of tokens).
Another version (on the right) only considers a 1-init window with 1.01$^k$
($k=1,2,3, \cdots$, round up to the nearest integer) tokens. The number of data points
is listed in column-6. The $\alpha$, $\beta$ columns are fitting values of the
linear and quadratic terms in Eq.(\ref{eq-q}), and $R^2_{adj}$ is the adjusted
$R^2$.
}
\end{table}

The results from both approaches (left and right halves of Table \ref{table2})
are more or less the same. The $\beta$ is always negative, with the exception
of {\sl Natural History} which is encyclopedia-like.  The median value of 
$\beta$ from the left side of Table \ref{table2} is $-0.025$, and that from 
the right side is $-0.021$.  The $\alpha$ values are all around 1. 
The median value of $\alpha$ from the left side of Table \ref{table2}
is $1.09$, and that from the right side is 1.03.  The fact of $\alpha\approx 1$
might seem to be a violation of the Heaps' law, where $\alpha$ is strictly
smaller than 1. However, the negative $\beta$ value plays the role of $\alpha <1$
to bend the token-type curve, not only in linear-linear scale, but in log-log scale as well. 
The intercept $c_0$ (see Eq.(\ref{eq-q})) of the regression is closely related to the value of $\alpha$:
whenever $\alpha >1$, the intercept is negative, and
when $\alpha <1$, the intercept is positive 
(result not shown).

\section{Results from  the random ball drawing model }

\subsection{ Curvature in log-log scale is identical to a ``pseudo-variance"}
\label{sec:pseudoV}

\indent

In the random ball drawing model (or simply urn model) 
of the token-type relationship, a token is considered
as a colored ball from a bag, and different colors mean different types. If the ball
is put back after drawing, it is the situation of drawing with replacement. The
probability of observing a color (word type) remains a constant independent of how
many balls have already been drawn from the bag. We use $p_i$ to denote this constant
probability of observing color-$i$, and Eq.(\ref{eq-w-replace}) can be rewritten as:
\begin{equation}
\label{eq-w-replace2}
E[V(T)]= V_{tot} - \sum_{i=1}^{V_{tot}} \left( 1-p_i \right)^T
\end{equation}
For a small number of tokens, the random ball model may not be correct, as
there are correlation between words in a sentence. However, we believe that
at the whole text level, when $T$ is large, the random ball model is accurate.

Using the binomial theorem, we have ($C(T,j)=T!/[j! (T-j)!]$ is the number of combinations
of choosing $j$ objects from a total of $T$ objects):
\begin{eqnarray}
E[V(T)] &=& V_{tot}- \sum_{i=1}^{V_{tot}} \sum_{j=0}^T  C(T,j) (-1)^j p_i^j \nonumber \\
 &=& V_{tot}- \sum_{i=1}^{V_{tot}} \left( 1 - Tp_i +\frac{T(T-1)}{2} p_i^2 + \cdots  + p_i^T\right)
 \nonumber \\
&=& \sum_{i=1}^{V_{tot}}\left(  T p_i - \frac{T(T-1)}{2} p_i^2 + \cdots  + p_i^T \right)
 \nonumber \\
&=& T - \frac{T(T-1)}{2} \sum_{i=1}^{V_{tot}} p_i^2 + \cdots  + \sum_{i=1}^{V_{tot}} p_iT
 \nonumber \\
&\equiv & T - \frac{T(T-1)}{2} M_2 + \frac{T(T-1)(T-2)}{6} M_3 + \cdots  + M_T
\end{eqnarray}
where in the last line, 
\begin{equation}
\label{eq-moment}
 M_k \equiv \sum_i p_i^k
\end{equation}
is the $k$-th moment of the word type probability distribution. $M_1=\sum_i p_i$ is equal to 1.
We would like to express the above formula as a function of $T$. Towards that, we use
the definition of Stirling number (of the first type):
\begin{equation}
(T)_n \equiv  T(T-1)(T-2) \cdots (T-n+1) \equiv \sum_{k=0}^n s_k(n) T^k
\end{equation}
For example, $(T)_2= -T +T^2$, therefore, $s_1(2)=-1$ and $s_2(2)=1$.
Using the Stirling numbers and combining the coefficients for $T$, $T^2$, etc, we have
\begin{eqnarray}
\label{eq-T}
E[V(T)] &=& \left(M_1- \frac{s_1(2)}{2} M_2 + \frac{s_1(3)}{6} M_3 + \cdots \right) T +
\left( \frac{s_2(2)}{2}M_2 + \frac{s_2(3)}{6}M_3 + \cdots \right) T^2) + \cdots
 \nonumber \\
&=& \sum_{k=1}^T \left( \sum_{j=k}^{T} \frac{s_j(k)}{j!} (-1)^jM_j \right) T^k
\equiv \sum_{k=1}^T \beta_k(T) T^k
\end{eqnarray}
where the shorthand notation:
\begin{equation}
\label{eq-beta}
 \beta_k(T) \equiv \sum_{j=k}^{T}\frac{ s_j(k) (-1)^jM_j}{j! } 
\end{equation}
is introduced for convenience.

How do we reconcile the token-type relationship Eq.(\ref{eq-T}) from the random ball
drawing model and the quadratic regression in log-log scale in Eq.(\ref{eq-q})?
Using the transformed variable in log-log scale: $y=\log(V)$, $x=\log(T)$, the
second derivative of Eq.(\ref{eq-q}) is twice the
regression coefficient for the quadratic term, because:, 
\begin{equation}
\label{eq-twice}
\frac{d^2 y}{dx^2}  = \frac{d^2}{dx^2} (c_0 + \alpha x + \beta x^2)= 2 \beta .
\end{equation} 
Now we calculate the second derivative of Eq.(\ref{eq-T}) from the model.

The first derivative of Eq.(\ref{eq-T}) in log-log scale (also known as 
``elasticity" in the literature \citep{marshall}) is:
\begin{equation}
\label{eq-1st-der}
\frac{dy}{dx} =  \frac{d}{dx} \log \left(\sum_{k=1}^T \beta_k(T) e^{xk} \right)
= \frac{ \sum_{k=1}^T \beta_k(T) k e^{xk} }{\sum_{k=1}^T \beta_k(T) e^{xk}}
\equiv \sum_{k=1}^T \pi_k(x) k
\end{equation}
where $\pi_k(x)$ (note $x=\log(T)$)  is defined as a normalized $\beta_k(T)$:
\begin{equation}
\label{eq-pi}
\pi_k(T) =
\pi_k(x) \equiv \frac{\beta_k(T) e^{xk}}{\sum_{k=1}^T \beta_k(T) e^{xk}} 
= \frac{\beta_k(T) T^k}{\sum_{k=1}^T \beta_k(T) T^k} 
=\frac{ \left( \sum_{j=k}^{T} \frac{s_j(k) (-1)^jM_j}{j!} \right) T^k}
{\sum_{k=1}^T \left( \sum_{j=k}^{T} \frac{s_j(k) (-1)^jM_j}{j!} \right) T^k}
\end{equation}
Since $\sum_k \pi_k(x)=1$, these might be considered as weights. However, they are
not proper weights, because some of them can be negative. If we call $\pi_k(x)$
a ``pseudo weight" (or ``signed weight"), the first 
derivative in Eq.(\ref{eq-1st-der}) is a ``pseudo mean" (also called  
``affine combination" as vs. the standard convex combination, e.g. \citep{bershad}).

The second derivative (which does not have a standard name, but can be called
curvature in log-log scale, elasticity of elasticity, etc.)  in Eq.(\ref{eq-T}) is:
\begin{eqnarray}
\label{eq-2nd-der}
\frac{d^2y}{dx^2} &=& \sum_{k=1}^T k \frac{d}{dx} \pi_k(x) =
\sum_{k=1}^T k \frac{d}{dx} \left( \frac{\beta_k(T) e^{xk}}{\sum_{k=1}^T \beta_k(T) e^{xk} } \right)
 \nonumber \\
&=&\sum_{k=1}^T k \left( 
\frac{\beta_k(T) k e^{xk}}{\sum_{k=1}^T \beta_k(T) e^{xk}}
- \frac{\beta_k(T) e^{xk} \sum_{k=1}^T \beta_k(T) k e^{xk} }{ (\sum_{k=1}^T \beta_k(T) e^{xk})^2} \right)
 \nonumber \\
&=& \sum_{k=1}^T k^2 \pi_k(x) - \left( \sum_{k=1}^T k \pi_k(x) \right)^2
 \nonumber \\
&=& \sum_{k=1}^T \left( k - \sum_{k=1}^T k \pi_k(x)\right)^2 \pi_k(x)
\end{eqnarray}
It is of a form of variance, but since $\pi_k(x)$ can be negative, we call it ``pseudo variance"
(other names could be ``generalized variance with signed weights", ``generalized second moment", etc).
Eq.(\ref{eq-2nd-der}) can then be summarized as: the curvature in the log-log scale
is identical to a pseudo-variance.

To recap  this subsection: the vocabulary size $V_{tot}$ does not appear in the final
formula; the dictionary word type distribution information is all summarized in the
moments $M_k$ ($k=2,3, \cdots T$); 
the first derivative in the log-log scale is the  pseudo-mean,
or $\alpha + 2 \beta \log(T)$ in quadratic regression Eq.(\ref{eq-q});
the second derivative in the log-log
scale is the pseudo-variance,  or 2$\beta$ if Eq.(\ref{eq-q}) is used.

\subsection{ Curvature in log-log scale is negative}
\label{sec:negative}

\indent

Even though we have derived the curvature in log-log scale from the random ball drawing
model, we still do not know if it is positive or negative. It belongs to the situation
of ``indefinite quadratic form". We first examine the situation with $T=3$ tokens,
and show later that the conclusion from $T=3$ is confirmed for larger $T$s.

It can be shown that the three ``pseudo weights" for $T=3$ are:
\begin{eqnarray}
\pi_1(3) &=& \frac{1+M_2/2+M_3/3}{1-M_2+M_3/3} 
\nonumber \\
\pi_2(3) &=& -\frac{ 3M_2/2+3M_3/2}{1-M_2+M_3/3}
\nonumber \\
\pi_3(3) &=& \frac{ 3M_3/2}{1-M_2+M_3/3}
\end{eqnarray}
The pseudo-variance is:
\begin{eqnarray}
\label{eq-T3-var}
<T^2>-<T>^2 &=& 
 (\pi_1+4\pi_2 +9\pi_3) -(\pi_1+2\pi_2 + 3\pi_3)^2 \nonumber \\
&=& \frac{1-5.5M_2 +7.83 M_3}{1-M_2+M_3/3} -\left( \frac{1-2.5M_2 + 1.83M_3}{1-M_2+M_3/3} \right)^2
\nonumber \\
&\approx & (1-5.5M_2) (1+M_2) - (1-5M_2) (1+2M_2) \nonumber \\
&\approx & (1-4.5M_2) - (1-3M_2)  \nonumber \\
& \approx & -1.5 M_2 <  0
\end{eqnarray}
The approximation is based on the observation that higher order moments are
always (order of magnitude) smaller, e.g. $M_3 \ll M_2$.

We have run a numeric simulation with the Zipf's law distribution of word frequency
$p_{(r)}\sim 1/r^{1.01}$ with $V_{tot}=5000$ words. The second and third moments are
$M_2=0.02128$, $M_3=0.00179$. Without ignoring the $M_3$ term, the Eq.(\ref{eq-T3-var})
is roughly $-1.19M_2$. So even though the qualitative conclusion of Eq.(\ref{eq-T3-var})
is correct, that the pseudo-variance is negative, the numerical estimation of 
the pseudo-variance may not be $-1.5M_2$  if the higher order moment $M_3$ is
ignored.

\subsection{ Weaker curvature as the Zipf's law exponent increases}
\label{sec:zipfexp}

\indent

Although Eq.(\ref{eq-2nd-der}) provides an analytic solution to the second 
derivative in log-log scale from the random ball drawing model, there are
still practical difficulties in its application. The first issue is that the
random ball drawing model assumes a fixed number of colors (fixed 
vocabulary size), then the ratio of the number of tokens and
the fixed number of types  $T/V_{tot}$ may affect the curvature. Intuitively,
if  $T \gg V_{tot}$, the curvature is 
more obvious than the case when $T \approx V_{tot}$,
thus the curvature may increase with $T$.
 
The second difficulty concerns the pseudo weights $\pi_k(T)$ themselves.
Since these are not required to be positive, but sum up to 1, the
magnitude of their oscillation between positive and negative can be
exponentially large (when $T$ is large). This creates a numeric instability
as any round off error could be magnified to make the result meaningless.
However, Eq.(\ref{eq-2nd-der}) can still be useful to investigate other
features if we limit the range of $T$ to be outside the numeric instability.

Here we show that the curvature of log-log token-type plot will
decrease (less negative) when the Zipf's law exponent increases. 
The Zipf's law states that the word-type frequencies, when ranked from
high to low, follow an inverse-power-law distribution: 
\begin{equation}
\label{eq-zipf}
p_{(r)} =   \frac{c}{r^a}, \hspace{0.2in} \mbox{$r=1,2,\cdots$}
\end{equation}
where the normalization factor $c= (\sum_r 1/r^a)^{-1}$ (the notation $(r)$
means $r$ is ranked). When $a$ is changed (suppose the 
vocabulary size $V_{tot}$ is fixed), the moments $M_k$ ($k=2,3,\cdots$) also change,
which subsequently affect $\beta_k$ (Eq.(\ref{eq-beta})), $\pi_k$ (Eq.(\ref{eq-pi})),
then the second derivative Eq.(\ref{eq-2nd-der}) and the quadratic
regression coefficient (which is half the second derivative, see Eq.(\ref{eq-twice})).

We have carried out a small-scale simulation result to compare the pseudo-variance
estimation of the quadratic regression coefficient, and the actual fitting of
the token-type relation from the random ball drawing model (Eq.(\ref{eq-w-replace2})).
The vocabulary size is assumed to be $V_{tot}=50$.
The choice of upper limit of $T$ in Eq.(\ref{eq-2nd-der}) is to ensure that
we do not encounter numeric instability. We chose two values, $T=50$ and $T=80$.
The choice for regression fitting, we choose the maximum window size $T$ so
that the $y$-axis does not reach the saturation level of $V_{tot}=50$. We chose the
maximum window size to be $T=400$.

Table \ref{table3} shows that when the Zipf's law's exponent $a$ increases
from 1.01 to 1.2, the second moment $M_2$ increases, whereas the absolute
value of the second derivatives
(curvature) of the log-log plot, according to the pseudo-variance calculation
from Eq.(\ref{eq-2nd-der}), decrease. The decrease is confirmed by the
fitting of the token-type plot from the random ball drawing without replacement model.
The pseudo-variance estimations of the curvature is slightly smaller than the
actually observed, because we limit the upper limit of the summation
in Eq.(\ref{eq-beta}) and Eq.(\ref{eq-pi}) to be small to avoid numeric instability.
But these can be a proof-of-concept that curvature in log-log token-type plot
can be reliably estimated.

\begin{table}[H]  
\begin{center}
\begin{tabular}{c|c|cc|c}
\hline
Zipf's $a$ & $M_2$ & pseudo-var/2 & pseudo-var/2 & $\beta$  \\
 &$V_{tot}=50$) & $T\le50$& $T\le 80$& (from $T$=20 to 402) \\
\hline
1.01 & 0.082 & $-$0.076 & $-$0.092 & $-$0.095 \\
1.05 & 0.09  & $-$0.073 & $-$0.087 & $-$0.0906 \\
1.1  & 0.101 & $-$0.067 & $-$0.081 & $-$0.0849 \\
1.15 & 0.113 & $-$0.062 & $-$0.075 & $-$0.0792 \\
1.2 & 0.126 & $-$0.057 & $-$0.071 & $-$0.0736 \\
\hline
\end{tabular}
\end{center}
\caption{ \label{table3}
Simulation of a random sampling from a 50-word dictionary. The word
frequency distribution follows a Zipf's law (Eq.(\ref{eq-zipf})) with
exponent $a$ (column-1). The resulting second moment $M_2$ (Eq.(\ref{eq-moment})) 
is listed in column-2. The theoretical predictions of the the quadratic coefficient
$\beta$ by Eq.(\ref{eq-2nd-der}) with upper limit of $T=50$ and $T=80$ are
in columns-3,4. The least-square estimation of $\beta$ from
fitting of the token-type relationship by the random word-type 
sampling Eq.(\ref{eq-w-replace2}) from $T=20$ to $T=402$ is shown in column-5.}
\end{table}

\section{Discussion}

\indent

Heaps-Herdan law was proposed to emphasize a deviation
from the linear relationship between type counts (vocabulary size)
and token counts (text length), by using a power-law function,
with data fitting done on a relative small scale. More recent
works pointed out that power-law relationship between types and
tokens, equivalent to a linear relationship between log(type)
and log(token), also does not fit the data on larger scales.
This work validates the second deviation (first deviation being
on linear token-type, and second deviation on linear log(type)-log(token))
both by analyzing real data and by using theoretical models.

The data analysis part of this paper should be less controversial. Although
there are different versions (variants) in plotting types vs. token
(evenly samplings data points in linear-token scale or log-token scale,
averaging data points in all moving windows or not averaging,
using the first window only or using all moving windows, etc.)
the conclusion, that quadratic regression fits the data better
than linear regression (in log-log scale) remains the same.

Our work builds upon previous publications that Heaps' law may not be 
exactly correct or not exactly power-laws \citep{egghe,lv2013,bochkarev,font}. 
However, most of these works focus on a relationship between Heaps' law and 
Zipf's law: if both laws are exactly true within a range, and the question
asked was whether there is a relation between the two power-law 
exponents \citep{serrano,lv2010,eliazar13,font,boytsov,mazzolini}.
Our current work provides a rigorous statistical test on the
part of text analysis, and provide a more careful analysis considering
more than one way of drawing the token-type plot as well as doing
regression.

Punctuations are usually not considered to be part of
the vocabulary. However, they may have their own role in the language
\citep{nunberg}. Recent work shows that if punctuations are included
as part of the vocabulary, they follow the Zipf's law trend extended
to even higher frequencies \citep{kulig}. In fact, punctuations
may behave like stopwords that occupy all the top ranking spots in 
a rank-frequency plot \citep{wli-stop}, and mostly play a structural role instead 
of a semantic one. However, due to the limited number of punctuations, 
they are not expected to contribute to the word-type growth in token-type plot.

On the theoretical side, we provide a formula for the curvature
(quadratic term correction). Previous analytic works are most in an integral
form and/or their linear regimes  \citep{eliazar11,font,boytsov}.
Our formula on the curvature in Sec 3.1 does not require Zipf's law
to be true, and can be used with any rank-frequency distributions. 
We recognize that the random ball-drawing model (urn sampling model)
is not realistic, which assumes that word selections occur independently. 
But, this independent sampling framework serves as a valuable null model for 
understanding vocabulary growth,  which has enough features to lead 
to our main result on the second-order correction 
to Heaps' law via the frequency moments of the distribution. In natural language,
words are interconnected through syntax, semantics, discourse structure,
and topic persistence \citep{drozdz}. The correlation between tokens is
more likely to be positive, therefore reducing the chance to introduce new
word types. We expect such a correlation may lead to a more negative curvature
in the log-log token-type plot. 

The token-type relationship in random ball drawing model is
known exactly, but that analytic expression is in a form of finite
series whose properties are basically unreadable. Previous
works (also including the version of sampling without replacement)
were mainly to extract first order information (Heaps' law exponent)
by different approximations \citep{font,boytsov}. The goal of our
analytic work is to extract the second order information from
this model, which is related to the coefficient of the quadratic
term in the regression.

The pseudo weights $\pi_k$ (Eq.(\ref{eq-pi})) have one disadvantage:
these are unstable from $k$ to $k+1$, and manage to
normalize to a unit sum only by cancel large positive values with
large negative ones. Therefore, Eq.(\ref{eq-2nd-der}) is not
yet practical to be used to estimate the $\beta$ in Eq.(\ref{eq-q})
for large vocabulary sizes and large token counts. However, for
smaller token counts and small vocabulary sizes, and for qualitative
argument, these analytic formula are very useful.

In Eq.(\ref{eq-T3-var}), we have shown for 3 tokens 
that  the pseudo-variance is negative regardless of word frequency distribution
-- Zipf's law or not, according to the urn model. However, a particular
distribution may curve more severely than other distributions. From
computer runs of the urn model, we observed that distributions whose
moments decrease more slowly (e.g. Zipf's law with larger exponent)
curve less than those whose moments decrease more rapidly (e.g., 
uniform distribution), when the vocabulary size is fixed. When the  vocabulary
size as well as the number of tokens increase, we still consistently
observe the negative curvature from the numerical simulation, therein
less severely. 

Although the quadratic regression in log(type)-log(token) plot
improves upon Heaps' law, it is still an approximation.
Since the coefficient $\beta$ in Eq.(\ref{eq-q}) is negative,
it is concave down, and will eventually change a positive
slope in log(type)-log(token) plot to zero slope, then to negative
(which should not be allowed). However, the turning point to be
reached by the quadratic function is far away: using the {\sl War
and Peace} regression result for example (Table \ref{table2}), the number of tokens
need to be $exp(\alpha/(2 |\beta|)) = exp(1.07/(2 \cdot 0.024)) = exp(22.29)= 4.8 \times 10^9 $,
to see the quadratic function break down. Apparently, it is hard
to reach this text length, which is equivalent to 8478 volumes
of {\sl War and Peace} itself.

The urn model of random drawing colored balls with replacement 
Eq.(\ref{eq-w-replace}) or Eq.(\ref{eq-w-replace2}) can also be written in 
the following approximation, assuming $p_i$'s to be small: 
\begin{equation}
\label{eq-w-replace3}
E[V(T)]= V_{tot} - \sum_{i=1}^{V_{tot}} e^{ T \log( 1-p_i )}
\approx  V_{tot} - \sum_{i=1}^{V_{tot}} e^{ -T p_i}
= \sum_{i=1}^{V_{tot}} (1- e^{ -T p_i}).
\end{equation}
It is called the Poisson approximation in \citep{baayen},
though the above derivation does not refer a use of Poisson distribution.
Although Eq.(\ref{eq-w-replace3}) is indeed simpler than
Eq.(\ref{eq-w-replace}) or Eq.(\ref{eq-w-replace2}), it is
simpler in $V$ vs. $T$, not in $\log(V)$ vs. $\log(T)$.
Further works are needed to find an analytic expression
that would make it possible to carry out a numerical plug-in
calculation of the curvature in log(type)-log(token) plot with
high precision. 

To summarize, we present convincing evidences that the Heaps' law,
though good as an approximation, does not and will not fit the
token-token relationship exactly. We propose a quadratic
correction in log(type)-log(token) plot and it improves
the data fitting over the Heaps' law.  Using the random ball
drawing model, which we argue should be an appropriate model
for studying token-type curvature, we have proved that the
curvature in log(type)-log(token) is equal to a pseudo-variance,
and giving evidence that this pseudo-variance should be negative.

\section{Data and Programs Used}
\label{sec:data}

\indent

Twenty eBooks, 
are downloaded from the Project Gutenberg
({\url{https://www.gutenberg.org/}). 
These 20 books are (ranked by the text length from short to long):
1. {\sl Discourse on the Method of Rightly Conducting One's Reason and of Seeking Truth 
in the Sciences} (by Ren\'{e} Descartes (1637), autobiographical writing translated 
from French, 23027 token),
2. {\sl Alice's Adventure in Wonderland} (by Lewis Carroll (1865), children's book, 26687 tokens), 
3. {\sl Hamlet} (by William Shakespeare (1599-1601), play, 31966 tokens), 
4.  {\sl Beowulf} (by an unknown author (8th to 11th centuries), old English poem, 36171 tokens), 
5.  {\sl Utopia}(by Saint Thomas More (1516), philosophic work translated from Latin, 42249 tokens),
6.  {\sl Great Gastby} (by F Scott Fitzgerald (1925), novel, 48661 tokens), 
7.  {\sl Frankenstein}(by Mary Shelley (1818), novel, 69632 tokens),
8.  {\sl Natural History of Pliny} (by Pliny the Elder (AD77), encyclopedia translated from Latin, 
Vol 1, 90675 tokens)
9.  {\sl Pride and Prejudice}(by Jane Austen (1813), novel, 122971 tokens),
10.
{\sl An Essay Concerning Humane Understanding} (by John Locke (1689), philosophic work, 143672 tokens),
11.
{\sl On the Origin of Species} (by Charles Darwin (1859), scientific literature, 151077 tokens), 
12.
{\sl Dracula} (by Bram Stoker (1897), horror novel, 162019 tokens),
13.
{\sl Leviathan} (by Thomas Hobbes (1651), social science book, 211782 tokens), 
14.
{\sl Moby Dick} (by Herman Melville (1851), novel, 212510 tokens), 
15.
{\sl Ulysses} (by James Joyce (1922), modernist novel,  265191 tokens), 
16.
{\sl Middlemarch}(by George Eliot (1871-1872), novel, 320171 tokens),
17.
{\sl Bleak House} (by Charles Dickens (1852-1853), novel, 357498 tokens),
18.
{\sl Don Quijote} (by Miguel de Cervantes Saavedra (1605-1615), novel translated from Spanish to English,
 376691 tokens), 
19.
{\sl An Inquiry into the Nature and Causes of the  Wealth of Nations} 
(by Adam Smith(1776), economic treatise, 380995 tokens), 
and 
20.
{\sl War and Peace} (by Leo Tolstoy (1869), novel translated from Russian to English, 566051 tokens).

Text manipulation was carried out in the $R$ statistical computational 
environment  (\url{https://www.r-project.org/}), both the basic functions
(e.g. {\sl lm()} for linear regression)
and from the {\sl tidyverse} package collection, including ggplot2, 
dplyr, tidyr, tidytext \citep{silge}.

\newpage

\begin{figure}[t]
 \begin{center}
 \includegraphics[width=0.9\textwidth]{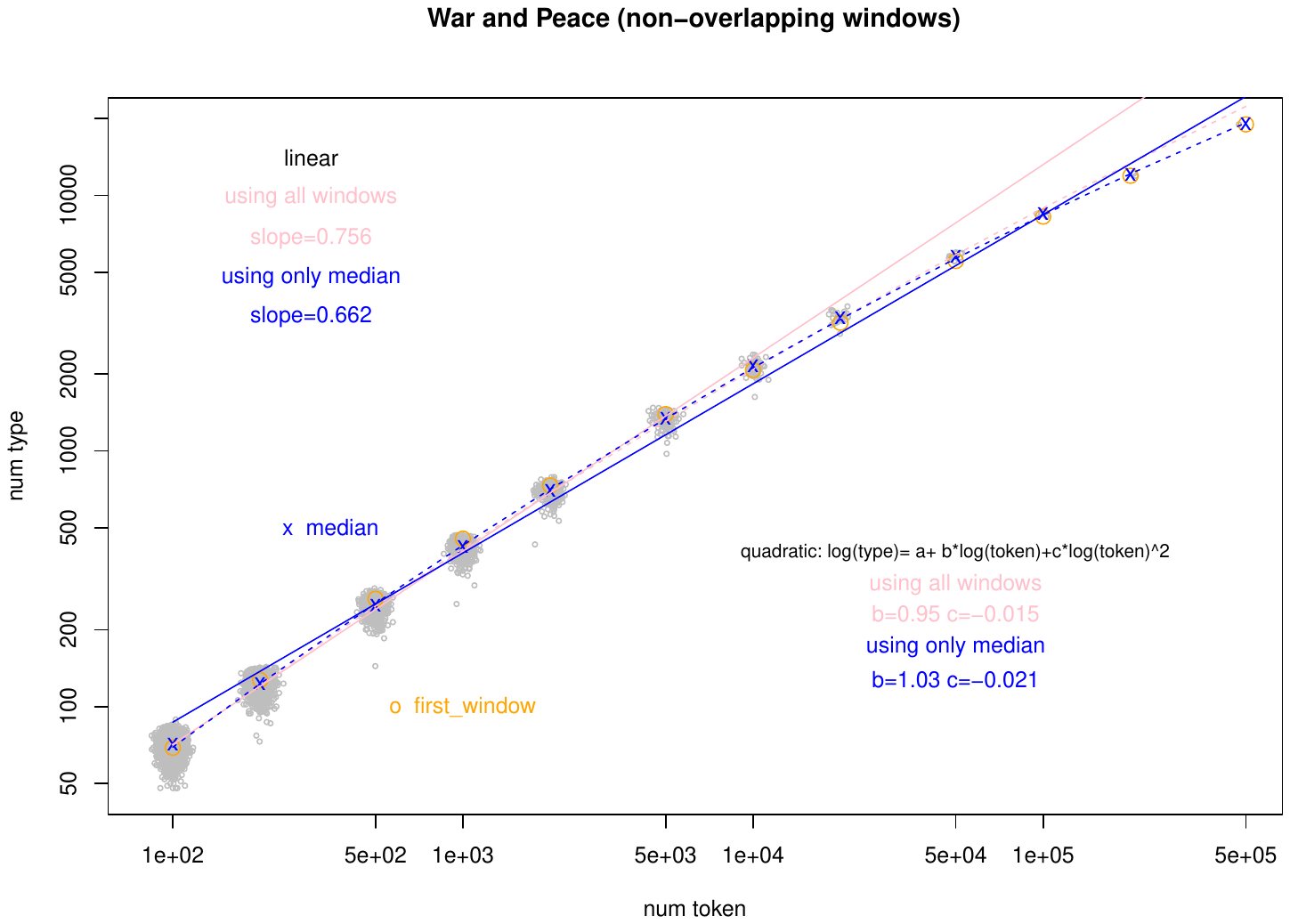}
 \end{center}
\caption{ \label{fig1}
Word-type ($y$-axis) vs. word token ($x$-axis) counts in an English translation
of {\sl War and Peace}, in log-log scale. the grey cloud around, e.g., $x=100$ and 
$y=70$, are word-type counts in (non-overlapping) moving window of 100 tokens
along the text (noise is added to $x=100$). Similar grey clouds are done similar
for moving windows of sizes 200, 500, 1000, $\cdots$ tokens. The last point at
$x=500,000$ more or less cover the whole book. The orange circles indicate the
word-type count in the first window. The blue crosses indicate  the median of word-type
counts with the same window size. Linear (solid line) and quadratic regressions 
(dashed line) on all windows (grey points, pink lines) and on medians (blue crosses,
blue lines) are shown.
}
\end{figure}

\begin{figure}[t]
 \begin{center}
 \includegraphics[width=1.0\textwidth]{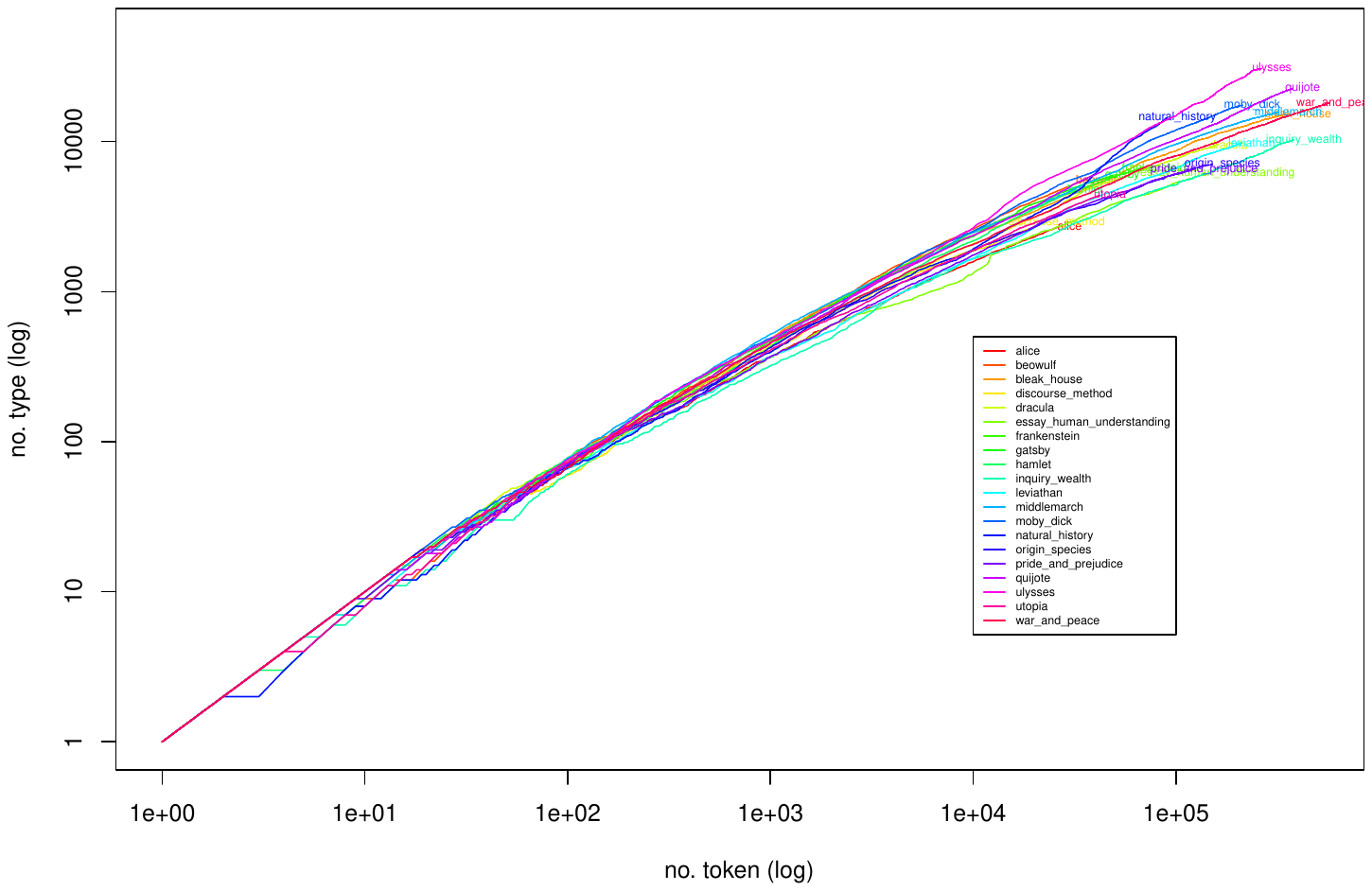}
 \end{center}
\caption{ \label{fig2}
Word-type ($y$-axis) vs. token ($x$-axis) counts for 20 books from
the Project Gutenberg, in log-log scale.
}
\end{figure}

\begin{figure}[t]
 \begin{center}
 \includegraphics[width=1.1\textwidth]{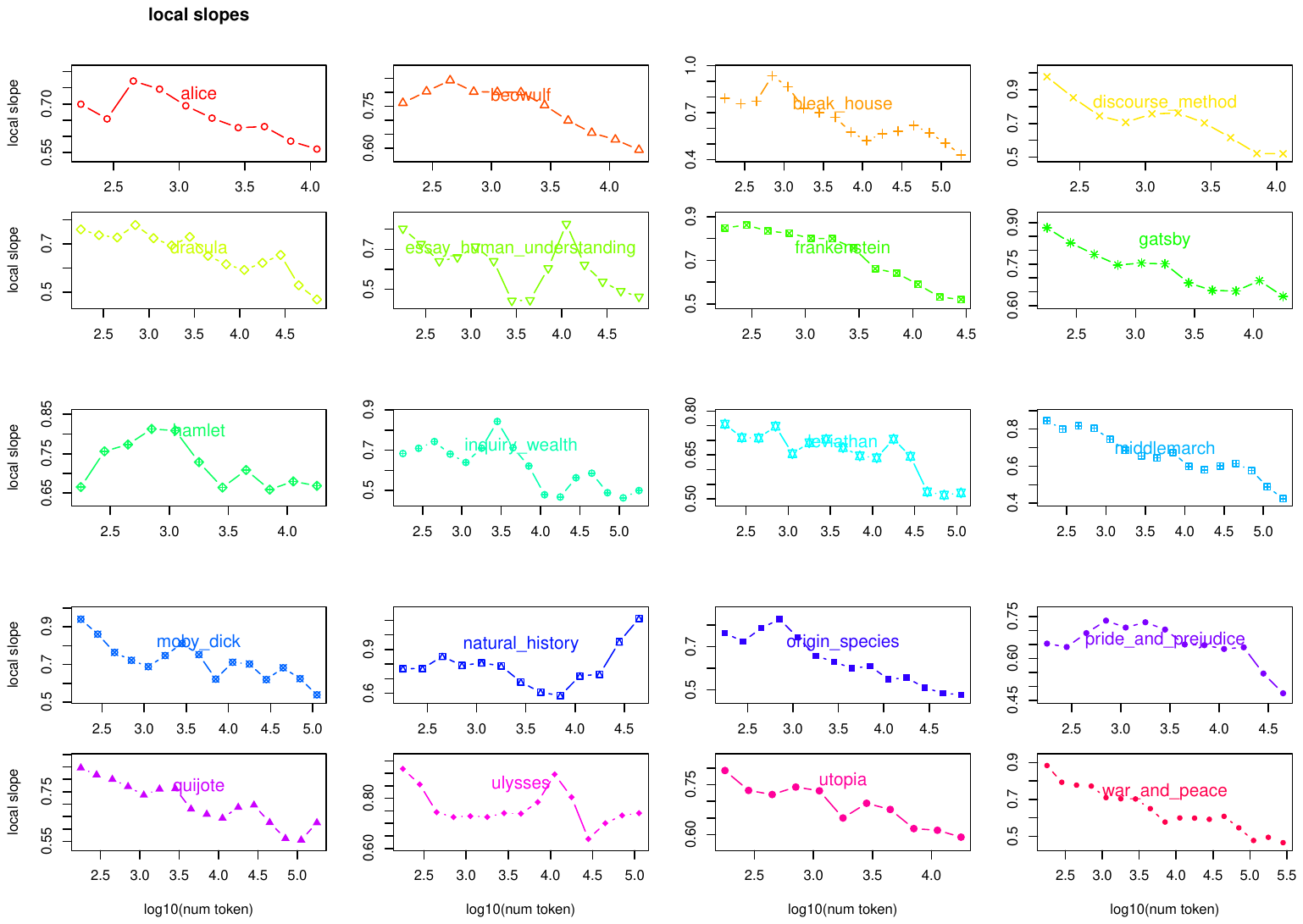}
 \end{center}
\caption{ \label{fig3}
Local slopes in $\log_{10}$(type) over $\log_{10}$(token) plots (also
called local elasticity) for the 20 books, as a function of $\log_{10}$)
token counts. 
}
\end{figure}


\newpage


\normalsize

\begin{thebibliography}{99}

\bibitem[Akaike,(1974)]{aic} 
H Akaike (1974),
A new look at the statistical model identification,
{\sl IEEE Trans. Automatic Control}, 19:716-723.

\bibitem[Altmann and Gerlach,(2016)]{altmann} 
EG Altmann and M Gerlach (2016),
Statistical laws in linguistics,
in {\sl Creativity and Universality in Language},
eds. MD Esposti, EG Altmann, F Pachet, pp 7-26 (Springer).

\bibitem[Baayen,(2001)]{baayen} 
RH Baayen (2001),
{\sl Word Frequency Distributions} (Kluwer Academic Publishers, Dordrecht).

\bibitem[Bernhardsson et al.,(2009)]{bernhardsson} 
S Bernhardsson,  LEC de Rocha, P Minnhagen (2009),
The meta book and size-dependent properties of written language,
{\sl New J. Phys.}, 11:123015.

\bibitem[Bershad, et al.,(2008)]{bershad} 
NJ Bershad, JCM Bermudez, JY Tourneret (2008),
An affine combination of two LMS adaptive filters -- transient mean-square analysis, 
{\sl IEEE Trans, Signal Proc.}, 56:1853-1864.

\bibitem[Bail\'{o}n-Moreno et al.,(2005)]{bailon} 
R Bail\'{o}n-Moreno, E Jurado-Alameda, R Ruiz-Na\~{n}os, JP Courtial (2005),
Bibliometric laws: Empirical flaws of fit,
{\sl Scientometrics}, 63:209-229.

\bibitem[Bochkarev et al.,(2014)]{bochkarev} 
VV Bochkarev, EY Lerner, AV Shevlyakova (2014),
Deviations in the Zipf and Heaps laws in natural languages,
{\sl J. Phys.: Conf. Series}, 490:012009.

\bibitem[Boytsov,(2017)]{boytsov} 
L Boytsov (2017),
A simple derivation of the Heap's law from the generalized Zipf's law,
{\sl arXiv} preprint, 1711.03066.

\bibitem[Chacoma and Zanette,(2020)]{chacoma} 
A Chacoma and DH Zanette (2020),
Heaps’ Law and Heaps functions in tagged texts: evidences of their
linguistic relevance,
{\sl Royal Soc. Open Sci.}, 7:200008. 


\bibitem[Davis,(2019)]{davis} 
V Davis (2019),
Types, tokens, and hapaxes: a new Heap's law,
{\sl Glottotheory}, 9:113-129.

\bibitem[D\d{e}bowski,(2025)]{debowski} 
\L D\d{e}boski (2025),
Corrections of Zipf's and Heaps' laws derived from hapax rate models,
{\sl J. Quant. Linguis.}, 32:128-165.

\bibitem[Dro\.{z}d\.{z} et al.,(2025)]{drozdz} 
S Dro\.{z}d\.{z}, J Kwapie\'{n}, T Stanisz (2025),
Editorial: complexity characteristics of natural language,
{\sl Entropy}, 28:98.

\bibitem[Egghe,(2007)]{egghe} 
L Egghe (2007),
Untangling Herdan’s law and Heaps’ law: mathematical and informetric arguments,
{\sl J. Am. Soc. Info. Sci. Tech.}, 58:702-709.

\bibitem[Eliazar,(2011)]{eliazar11} 
I Elizar (2011),
The growth statistics of Zipfian ensembles: Beyond Heaps’ law,
{\sl Phys. A}, 390:3189-3203. 

\bibitem[Eliazar and Cohen,(2013)]{eliazar13} 
II Elizar and MH Cohen (2013),
Power-law connections: From Zipf to Heaps and beyond,
{\sl Ann. Phys.}, 332:56-74.

\bibitem[Ezekiel,(1930)]{ezekiel} 
M Ezekiel (1930),
{\sl Methods of Correlation Analysis} (Wiley).

\bibitem[Font-Clos and Corral,(2015)]{font} 
F Font-Clos and \'{A} Corral (2015),
Log-Log convexity of type-token growth in Zipf’s systems,
{\sl Phys. Rev. Lett.}, 114:238701.

\bibitem[Fontanelli et al.,(2022)]{oscar-brf} 
O Fontanelli, P Miramontes, R Mansilla, G Cocho, W Li (2022),
Beta rank function: a smooth double-Pareto-like distribution,
{\sl Comm. Stat. -- Theory and Meth.}, 51:3645-3668.

\bibitem[Fontanelli et al.,(2016)]{oscar-lavalette} 
O Fontanelli, P Miramontes, Y Yang, G Cocho, W Li (2016),
Beyond Zipf’s Law: the Lavalette rank function and its properties,
{\sl PLoS ONE}, 11:e0163241.

\bibitem[Frappat et al.,(2003)]{frappat} 
L Frappat, C Minichini, A Sciarrino, P Sorba (2003),
Universality and Shannon entropy of codon usage,
{\sl Phys. Rev. E}, 68:061910. 


\bibitem[Heaps,(1978)]{heaps} 
HS Heaps (1978),
{\sl Information Retrieval: Computational and Theoretical Aspects}
(Academic Press, New York, USA).

\bibitem[Heck Jr. etc.,(1975)]{heck} 
KL Heck Jr. G van Belle, D Simberloff (1975),
Explicit calculation of the rarefaction diversity measurement and the determination of 
sufficient sample size,
{\sl Ecology}, 56:1459-1461.

\bibitem[Herdan,(1960)]{tt-herdan} 
G Herdan (1960),
{\sl Type-token Mathematics: A Textbook of Mathematical Linguistics}
(Mouton, The Hague, Netherlands).

\bibitem[Hurlbert,(1971)]{hurlbert} 
SH Hurlbert (1971),
The nonconcept of species diversity: a critique and alternative parameters,
{\sl Ecology}, 52:577-586.

\bibitem[Kulig et al.,(2017)]{kulig} 
A Kulig, J Kwapie\'{n}, T Stanisz, S Dro\.{z}d\.{z} (2017),
In narrative texts punctuation marks obey the same statistics as words,
{\sl Info. Sci.}, 375:98-113.

\bibitem[Kretzschmar Jr.,(2015)]{kret} 
WA Kretzschmar Jr. (2015),
{\sl Language and Complex Systems} (Cambridge University Press).

\bibitem[Larsen-Freeman and Cameron,(2008)]{larsen} 
D Larsen-Freemand and L Cameron (2008),
{\sl Complex Systems and Applied Linguistics} (Oxford University Press).

\bibitem[Li et al.,(2024)]{wli24} 
W Li, Y Almirantis, A Provata (2024),
Range-limited Heaps' law for functional DNA words in the human genome,
{\sl J. Theo. Biol.}, 592:111878.

\bibitem[Li and Fontanelli.,(2026)]{wli-stop} 
W Li and O Fontanelli (2026),
Non-Zipfian distribution of stopwords and subset selection models,
{\sl arXiv} preprint, arXiv:2603.04691.

\bibitem[Li and Miramontes,(2011)]{wli-letter} 
W Li and P Miramontes (2011),
Fitting ranked English and Spanish letter frequency distribution 
in US and Mexican presidential speeches,
{\sl J. Quant. Linguist.}, 18:359-380.

\bibitem[Li et al.,(2010)]{wli-entropy} 
W Li, P Miramontes, G Cocho (2010),
Fitting ranked linguistic data with two-parameter functions,
{\sl Entropy}, 12:1743-1764.

\bibitem[Liberman et al.,(2007)]{lieberman} 
E Lieberman, JB Michel, J Jackson, T Tang, MA Nowak (2007),
Quantifying the evolutionary dynamics of language,
{\sl Nature}, 449:713-716.

\bibitem[L\"{u} et al.,(2010)]{lv2010} 
L L\"{u}, ZK Zhang, T Zhao (2010),
Zipf's Law leads to Heaps' law: analyzing their relation in finite-size systems,
{\sl PLoS ONE}, 5:e14139.

\bibitem[L\"{u} et al.,(2013)]{lv2013} 
L L\"{u}, ZK Zhang, T Zhou (2013),
Deviation of Zipf's and Heaps' Laws in human languages with limited dictionary sizes,
{\sl Sci. Rep.}, 3:1082.

\bibitem[Marshall,(1890)]{marshall} 
A Marshall (1890),
{\sl Principles of Economics} (Macmilan, London, UK).

\bibitem[Mazzolini et al.,(2018)]{mazzolini} 
A Mazzolini, J Grilli, E De Lazzari, M Osella, MC Lagomarsino, M Gherardi (2018),
Zipf and Heaps laws from dependency structures in component systems,
{\sl Phys. Rev.E}, 98:012315.


\bibitem[Mehri and Jamaati,(2017)]{mehri} 
A Mehri and M Jamaati (2017),
Variation of Zipf's exponent in one hundred live languages: 
a study of the Holy Bible translations,
{\sl Phys. Lett. A}, 381:2471-2477.


\bibitem[Mili\v{c}ka,(2009)]{milicka09} 
J Mili\v{c}ka (2009)
Type-token \& hapax-token relation: a combinatorial model,
{\sl Glottotheory}, 1:99-110.


\bibitem[Moreno-S\'{a}nchez et al.,(2016)]{moreno} 
I Moreno-S\'{a}nchez, F Font-Clos, \'{A} Corral (2016),
Large-scale analysis of Zipf’s law in English texts,
{\sl PLoS ONE}, 11:e0147073.

\bibitem[Nunberg,(1983)]{nunberg} 
G Nunberg (1983),
{\sl The Linguistics of Punctuation} (CSLI, Stanford University, Stanford, CA, USA).

\bibitem[Petersen et al.,(2016)]{petersen} 
C Petersen, JG Simonsen, C Lioma (2016),
Power law distributions in information retrieval,
{\sl ACM Trans. Inf. Sys.}, 34:1-37. 

\bibitem[Piantadosi,(2014)]{piantodosi} 
ST Piantadosi (2014),
Zipf’s word frequency law in natural language: A critical review
and future directions,
{\sl Psychonomic. Bull. Rev.}, 21:1112-1130.

\bibitem[Semple et al.,(2022)]{semple} 
S Semple, R Ferrer-i-Cancho, ML Gustison (2022),
Linguistic laws in biology, 
{\sl Trends Ecol. Evol.}, 37:P53-P66.

\bibitem[Serrano et al.,(2009)]{serrano} 
MA Serrano, A Flammini, F Menczer (2009),
Modeling statistical properties of written text,
{\sl PLoS ONE}, 4:e5372.

\bibitem[Silge and Robinson,(2017)]{silge} 
J Silge and D Robinson (2017),
{\sl Text Mining with R: A Tidy Approach} (O'Reilly). 

\bibitem[Smith et al.,(2003)]{smith} 
K Smith, H Brighton, S Kirby (2003),
Complex systems in language evolution: the cultural emergence
of compositional structure,
{\sl Adv. Complex Sys.}, 6:537-558.

\bibitem[Stanisz et al.,(2024)]{stanisz} 
T Stanisz, S Drozdz, J Kwapie\'{n} (2024),
Complex systems approach to natural language,
{\sl Phys. Rep.}, 1053:1-84.

\bibitem[Torre et al.,(2019)]{torre} 
IG Torre, B Luque, L Lacasa, CT Kello, A Hern\'{a}ndez-Fern\'{a}ndez (2019),
On the physical origin of linguistic laws and lognormality in speech,
{\sl Royal Soc. Open Sci.}, 6:191023.

\bibitem[Zipf,(1935)]{zipf35} 
GK Zipf (1935),
{\sl The Psycho-Biology of Languages} (Houghton-Mifflin, Boston, MA).


 
 

\end{thebibliography}
\end{document}